\documentclass[runningheads]{llncs}
\newcommand{\UE}{U_{\mathrm{E}}}
\usepackage{eccv}
\usepackage{eccvabbrv}

\usepackage{algorithm}
\usepackage{algpseudocode}
\usepackage{graphicx}
\usepackage{booktabs}
\usepackage[accsupp]{axessibility}

\usepackage{eso-pic}
\usepackage{graphicx}

\usepackage{hyperref}
\newcommand{\Cnom}{C_{\mathrm{nom}}}
\hypersetup{
    pdftitle={Worst-Case Hidden-Vehicle Trajectory Search in Spatiotemporal Occlusion Regions},
    pdfauthor={Ruichen Tan, Zengxiang Lei, Satish Ukkusuri},
    hidelinks
}

\begin{document}

\title{Worst-Case Hidden-Vehicle Trajectory Search in Spatiotemporal Occlusion Regions}
\titlerunning{Worst-Case Hidden-Vehicle Trajectory Search}

\author{Ruichen Tan\inst{1} \and
Zengxiang Lei\inst{1} \and
Satish Ukkusuri\inst{1}\thanks{Corresponding author.}}
\authorrunning{R. Tan et al.}

\institute{Lyles School of Civil and Construction Engineering, Purdue University,\\
West Lafayette, IN 47907, USA\\
\email{\{tan479,lei67,sukkusur\}@purdue.edu}}

\AddToShipoutPictureFG*{%
  \AtPageLowerLeft{%
    \put(\LenToUnit{12pt},\LenToUnit{0.5\paperheight}){%
      \makebox(0,0){%
        \rotatebox{90}{%
          \footnotesize
          Accepted at the ECCV 2026 Workshop on Safe and Defensive Autonomous Driving (SDAD).
          Best Paper Award.
        }%
      }%
    }%
  }%
}

\maketitle


\begin{abstract}
Occlusion creates fundamental uncertainty in autonomous driving. Existing methods often propagate frame-wise hypotheses or optimize ego behavior against prescribed hidden-agent predictions, leaving the worst history-consistent interaction unexplored. We introduce History-Conditioned Minimax Trajectory Search (HC-MTS), which combines temporal occlusion reasoning with response-aware search. First, HC-MTS constructs finite hidden-state modes, each certified by a backward witness satisfying multi-frame visibility, occupancy, semantic-map support, and class-specific kinematic constraints. It then solves a bilevel minimax problem: an inner finite oracle maximizes the ego driving score over destination attainment and ride comfort, while the outer search selects the legal hidden-vehicle trajectory that minimizes this best-response value. Across eight Waymo Open Motion Dataset scenarios, increasing the visibility-memory horizon from $K=1$ to $K=20$ reduces the mean per-scenario vehicle, pedestrian, and total retained hidden-seed counts by $18.12\%$, $21.67\%$, and $18.45\%$, respectively. HC-MTS identifies six avoidable counterexamples, while no legal collision-producing attacker is found in the remaining two scenes within the finite search budget.

\end{abstract}

\section{Introduction}

Occlusion introduces a fundamental source of uncertainty in autonomous driving. Vehicles and pedestrians concealed by traffic, roadside structures, or road geometry may be absent from the current perception output while remaining capable of entering the ego vehicle's path within the planning horizon. Treating occluded space as free is therefore unsafe, whereas assuming that every blind location is occupied leads to unnecessarily conservative behavior. Effective occlusion reasoning must distinguish physically plausible hidden participants from hypotheses that are incompatible with the ego vehicle's observation history. In particular, a state that is geometrically feasible within the current blind area should be discarded when no admissible trajectory could have reached that state without passing through previously visible and unoccupied space \cite{yu2019occlusion,orzechowski2018tackling,koschi2021set,wang2021reasoning,sanchez2022foresee}.

A common class of occlusion-aware methods initializes hypothetical participants from the current field of view and propagates their possible states for risk estimation, safety verification, or motion planning \cite{yu2019occlusion,orzechowski2018tackling,koschi2021set,nager2019shadows,moller2024overcoming}. Probabilistic and learning-based approaches estimate likely hidden occupancy or motion, while POMDP, game-theoretic, and contingency-based methods optimize ego behavior under partial observability \cite{hubmann2019pomdp,zhang2021occlusiongame,christianos2022bivo,qiu2023occluded}. Although these approaches address hidden-state estimation and decision-making under uncertainty, safety-critical evaluation requires jointly reasoning about both components. A hidden trajectory that collides with the nominal ego plan may impose little actual risk if the ego vehicle can avoid it without substantially sacrificing progress or comfort. Conversely, a less obvious trajectory may remain highly disruptive even after the ego vehicle selects its strongest available response. Identifying the most consequential hidden trajectory therefore requires evaluating each candidate against the ego vehicle's optimized response rather than against a fixed nominal plan.

We introduce \emph{History-Conditioned Minimax Trajectory Search} (HC-MTS), a two-stage framework combining temporal occlusion reasoning with response-aware worst-case trajectory discovery. In Stage~I, HC-MTS samples class-specific current speeds and backward control hypotheses for candidate states within the current blind area. A candidate is retained as a finite hidden-state mode only when it admits at least one backward witness satisfying multi-frame visibility, observed-occupancy, semantic-map, and class-specific kinematic constraints over the selected memory window. All memory horizons are evaluated from the same sampled modes using a prefix-validity test. Consequently, extending the observation history can eliminate unsupported hypotheses but cannot introduce new ones.

In Stage~II, the certified modes initialize a collision-conditioned bilevel minimax search. Candidate hidden trajectories must satisfy forward dynamics, geometry, semantic-map, occupancy, and legality constraints. To concentrate the finite search on safety-relevant interactions, each proposal must also collide with the nominal ego plan. This collision condition serves only as a proposal mechanism; the final severity of a hidden trajectory is determined after the ego vehicle is allowed to respond. Let $\mathcal{T}_K$ denote the set of legal hidden-vehicle trajectories initialized from modes certified over a $K$-frame history, and let $\mathcal{R}_e(\tau_h)$ denote the finite set of ego responses considered for a hidden trajectory $\tau_h$. HC-MTS approximates
\begin{equation}
\tau_h^\star \in \arg\min_{\tau_h \in \mathcal{T}_K}
\max_{\tau_e \in \mathcal{R}_e(\tau_h)}
J(\tau_e,\tau_h),
\end{equation}
where $J$ measures task-level driving performance through destination attainment and ride comfort. For each candidate hidden trajectory, the inner response oracle identifies the highest-scoring ego maneuver available within its finite response set. The outer search then seeks a legal hidden trajectory that minimizes this best-response value. The refinement process consequently prioritizes trajectories that remain damaging after accounting for the strongest ego response found by the oracle. The implemented method is a budgeted finite-oracle approximation using collision-conditioned proposals and a Gaussian elite refinement; it is not presented as a converged solution to a continuous bilevel optimization problem.

We evaluate HC-MTS on eight vectorized scenarios from the Waymo Open Motion Dataset (WOMD) \cite{ettinger2021womd}. The experiments examine how multi-frame visibility memory contracts the set of admissible hidden states and how the minimax search differentiates nominal-plan collisions that can be resolved by an ego response from counterexamples that remain unresolved under the current finite oracle. Together, these evaluations demonstrate the importance of combining temporal evidence with response-aware adversarial search when assessing risks created by occluded traffic participants.
Our contributions are threefold:
\begin{enumerate}
    \item We formulate hidden-state inference over multiple frames as a finite history-certification problem combining visibility, observed occupancy, semantic-map, and class-specific kinematic constraints. Every retained mode carries an explicit backward witness, and the retained set contracts monotonically with the memory horizon.
    \item We formulate hidden-trajectory discovery as a minimax problem: the ego response oracle maximizes task-level driving performance, while the adversarial search minimizes the resulting best-response score over legal history-certified trajectories.
    \item We provide an exact interface between inference and search: counterexample generation starts from the certified current modes without resampling their states or using hidden-agent future labels. Experiments on WOMD quantify both the contraction from temporal evidence and the counterexamples exposed by minimax search.
\end{enumerate}

\section{Related Work}

\subsection{History-Aware Reasoning about Occluded Participants}

Occlusion-aware methods represent unseen participants using sampled hypotheses, reachable occupancy sets, phantom agents, or learned probability distributions. Geometry- and set-based approaches initialize possible agents from the current field of view and propagate their states for risk estimation, safety verification, or motion planning \cite{yu2019occlusion,orzechowski2018tackling,koschi2021set,nager2019shadows,moller2024overcoming}. Learning-based methods instead infer likely hidden occupancy and motion from traffic data \cite{mahjourian2022occupancy,ange2024sceneinformer,christianos2022bivo}. Although these methods capture plausible future hazards, estimates derived primarily from the current blind area may retain states that could not have remained hidden throughout earlier observations. Temporal approaches address this limitation by tracking occluded regions or sequentially removing hidden states that conflict with newly available visibility evidence \cite{wang2021reasoning,sanchez2022foresee,nyberg2022evaluating,moller2025shadows}. HC-MTS builds on this temporal principle through an explicit history certificate: every retained current mode must admit a backward witness satisfying multi-frame visibility, observed-occupancy, semantic-map, and class-specific kinematic constraints. The certified modes then serve directly as initial conditions for worst-case trajectory search rather than only as occupancy bounds or inputs to an ego planner.

\subsection{Response-Aware Planning under Occlusion}

Partially observable, information-aware, and game-theoretic planners account for uncertainty when selecting ego behavior. Hubmann et al. incorporate potentially occluded vehicles and future visibility into a POMDP maneuver planner \cite{hubmann2019pomdp}, while Gilhuly et al. optimize trajectories for both safety and information acquisition \cite{gilhuly2022looking}. Zhang and Fisac formulate occlusion-aware driving as a hybrid dynamic game that accounts for adversarial hidden behavior and the ego vehicle's future response \cite{zhang2021occlusiongame}, and Qiu and Fridovich-Keil infer occluded-agent behavior within a receding-horizon contingency-game framework \cite{qiu2023occluded}. These methods primarily seek safe or informative ego policies under uncertainty. HC-MTS instead treats the hidden trajectory as the outer decision variable and evaluates its severity only after optimizing the ego response. Specifically, the inner finite oracle maximizes a task-level driving score based on destination attainment and comfort, while the outer search identifies a legal, history-certified hidden trajectory that minimizes this best-response value. The resulting output is therefore an interpretable worst-case hidden trajectory together with the strongest ego response found against it, rather than only a belief-conditioned ego action or binary safety guarantee.

\subsection{Safety-Critical Scenario and Trajectory Generation}

Safety-critical scenario-generation methods search for rare interactions that expose failures more efficiently than naturalistic sampling. Adaptive Stress Testing optimizes stochastic environment disturbances \cite{koren2018adaptive}; AdvSim perturbs actor trajectories while preserving physical plausibility \cite{wang2021advsim}; STRIVE searches the latent space of a learned traffic model for realistic collision-inducing scenes \cite{rempe2022strive}; and KING differentiates through a kinematic proxy to modify surrounding traffic adversarially \cite{hanselmann2022king}. Recent work also combines partial-observability risk estimation with generative models for adversarial scenario synthesis \cite{jia2026riskmap}. These approaches generally perturb visible actors, learned scene representations, or simulator parameters. HC-MTS addresses a different source of risk: an actor that may never have been observed. Its current state must first be supported by a feasible hidden history, and its future trajectory must satisfy dynamic, geometric, occupancy, map, and legality constraints. Furthermore, collision with the nominal ego plan is used only to generate critical candidates; final severity is determined by the optimized ego-response score. HC-MTS therefore extends safety-critical trajectory generation to history-conditioned hidden participants and evaluates adversariality through a response-aware minimax objective.

\section{Methodology}
\label{sec:method}

\subsection{HC-MTS Overview}
\label{sec:method_overview}
\begin{figure}[t]
    \centering
    \includegraphics[width=0.98\linewidth]{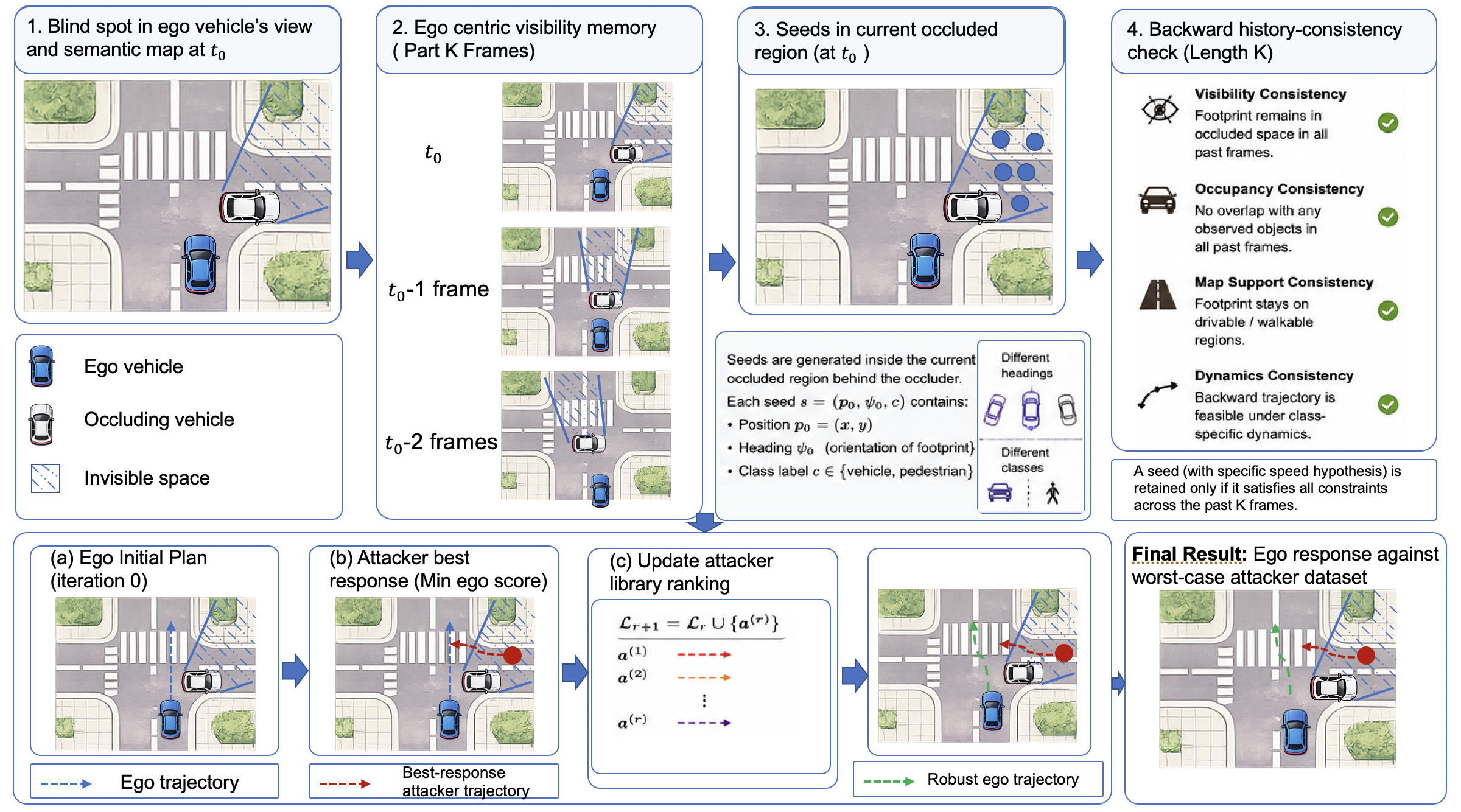}
     \caption{Overview of \emph{History-Conditioned Minimax Trajectory Search} (HC-MTS). Stage~I certifies current hidden-agent modes using backward witnesses that satisfy multi-frame visibility, observed-occupancy, semantic-map, and class-specific kinematic constraints. Stage~II initializes legal forward trajectories from the certified modes and performs a finite response-aware minimax search: an inner oracle maximizes the ego driving score, while the outer search seeks the hidden trajectory that minimizes this best-response value. Collision with the current ego plan is used to generate critical proposals, not to define their final severity.}
    \label{fig:overview}
\end{figure}

\emph{History-Conditioned Minimax Trajectory Search} (HC-MTS) identifies a history-consistent hidden trajectory that remains most detrimental after the ego vehicle is allowed to respond. At the current time $t_0$, HC-MTS receives the ego state, a vectorized semantic map, and observed road-user tracks from the current frame and the preceding $K$ frames. All scene elements are expressed in an ego-fixed bird's-eye-view frame anchored at $t_0$, with the ego vehicle at the origin, the $+x$ axis pointing forward, and the $+y$ axis pointing left.

HC-MTS consists of two stages. Stage~I performs \emph{history-conditioned hidden-mode inference}. It samples candidate current states in the blind area, augments them with class-specific speed and backward-control hypotheses, and retains only modes supported by at least one feasible backward witness through the observation history. Stage~II performs \emph{response-aware minimax trajectory search}. Each certified mode initializes legal forward hidden trajectories. For every candidate hidden trajectory, a finite ego-response oracle searches for the response that maximizes task-level driving performance. The outer search then favors hidden trajectories that reduce this optimized ego score. Figure \ref{fig:overview} separates feasibility from adversarial search: Stage I identifies history-consistent hidden modes, and Stage II optimizes their future trajectories for worst-case interaction with the ego vehicle.

\subsection{Problem Definition}
\label{sec:problem}

For each frame $t$, let $\Omega_t$ denote the field of regard, let $\mathcal B_t^{\mathrm{obs}}$ be the union of observed road-user footprints, and let $\mathcal F_t\subseteq\Omega_t$ denote the \emph{visible-free region}: space in which a road user would be expected to have been observed under the adopted geometric visibility model and that is not occupied by a detected participant. The unresolved non-visible region is
\begin{equation}
    \mathcal O_t
    =
    \Omega_t\setminus
    \left(\mathcal F_t\cup\mathcal B_t^{\mathrm{obs}}\right).
    \label{eq:occluded}
\end{equation}
Thus, $\mathcal F_t$, $\mathcal B_t^{\mathrm{obs}}$, and $\mathcal O_t$ distinguish observed free space, observed occupied space, and space whose occupancy remains unresolved. A hidden participant cannot intersect either of the first two sets, but it may occupy $\mathcal O_t$. 

Given the observation history from $t_0-K$ to $t_0$, Stage~I of HC-MTS returns a finite set of history-certified hidden-state modes
\begin{equation}
    \mathcal M_K
    =
    \left\{
        m\;\middle|\;\mathcal C_K(m)=1
    \right\},
    \label{eq:mk_definition}
\end{equation}
where $\mathcal C_K$ is the $K$-frame consistency predicate defined in Section~\ref{sec:history_inference}. Each $m\in\mathcal M_K$ contains an exact current hidden state and a backward witness demonstrating that the participant could have reached that state without violating the available visibility, occupancy, map, or kinematic evidence.

Let $\mathcal T(m)$ denote the set of forward hidden trajectories initialized from mode $m$ that satisfy the class-specific dynamics, footprint, semantic-map, observed-occupancy, and traffic-legality constraints. The history-conditioned admissible trajectory set is
\begin{equation}
    \mathcal T_K
    =
    \bigcup_{m\in\mathcal M_K}\mathcal T(m).
    \label{eq:admissible_hidden_trajectories}
\end{equation}


We make the ego-response objective explicit in a parameterized form. Let
$\tau_{\mathrm{ref}}$ denote the logged ego future, which is used only as a
reference trajectory. The nominal ego cost is defined as
\begin{align}
\Cnom(\tau_{\mathrm E};\boldsymbol{\theta}_{\mathrm E})
={}&
w_{\mathrm g}e_{\mathrm{goal}}^2
+w_{\mathrm p}e_{\mathrm{prog}}^2
+w_{\mathrm t}e_{\mathrm{time}}
+w_{\mathrm r}e_{\mathrm{route}}
\notag\\
&+
w_{\mathrm a}\,\overline{a^2}
+w_{\kappa}\,\overline{\kappa^2}
+w_{\mathrm s}\,\overline{(\Delta a)^2},
\label{eq:nominal-cost}
\end{align}
where
$\boldsymbol{\theta}_{\mathrm E}
=\{w_{\mathrm g},w_{\mathrm p},w_{\mathrm t},w_{\mathrm r},
w_{\mathrm a},w_{\kappa},w_{\mathrm s}\}$
contains nonnegative weighting parameters. The first four terms measure
goal-reaching, progress, time-aligned reference deviation, and route
deviation, respectively, while the remaining terms regularize control
effort and smoothness.

The ego utility is then defined as

\begin{equation}
\UE(\tau_{\mathrm E},\tau_{\mathrm A})=
\begin{cases}
-\infty,
& \text{if contact occurs},\\[1mm]
U_{\max}
-s_{\mathrm U}^{-1}
\left[
\Cnom(\tau_{\mathrm E};\boldsymbol{\theta}_{\mathrm E})
+\psi(d_{\min})
\right],
& \text{otherwise}.
\end{cases}
\label{eq:ego-score}
\end{equation}

where $d_{\min}$ is the minimum time-aligned oriented-box gap,
$w_{\mathrm d}>0$ controls the clearance penalty,
$\sigma_{\mathrm d}>0$ determines its spatial decay,
$s_{\mathrm U}>0$ is a normalization factor, and $U_{\max}$ denotes the
maximum collision-free utility. This formulation prioritizes collision
avoidance while favoring efficient, smooth, and well-separated ego
responses.

For a fixed hidden trajectory $\boldsymbol\tau^{A}\in\mathcal T_K$, let $\mathcal R_E(\boldsymbol\tau^{A})$ denote the ego responses available to the response oracle. We set
$J(\tau_{\mathrm E},\tau_{\mathrm A})\equiv
\UE(\tau_{\mathrm E},\tau_{\mathrm A})$. The best-response value of a hidden trajectory is
\begin{equation}
    V_E\!\left(\boldsymbol\tau^{A}\right)
    =
    \max_{\boldsymbol\tau^{E}\in\mathcal R_E(\boldsymbol\tau^{A})}
    J\!\left(\boldsymbol\tau^{E},\boldsymbol\tau^{A}\right).
    \label{eq:residual_ego_utility}
\end{equation}
HC-MTS targets the hidden trajectory that minimizes the ego vehicle's best achievable score:
\begin{equation}
    \boldsymbol\tau^{A*}
    \in
    \arg\min_{\boldsymbol\tau^{A}\in\mathcal T_K}
    \max_{\boldsymbol\tau^{E}\in\mathcal R_E(\boldsymbol\tau^{A})}
    J\!\left(\boldsymbol\tau^{E},\boldsymbol\tau^{A}\right)
    =
    \arg\min_{\boldsymbol\tau^{A}\in\mathcal T_K}
    V_E\!\left(\boldsymbol\tau^{A}\right).
    \label{eq:response_aware_attacker}
\end{equation}

\subsection{Ego-Centric Occlusion Representation}
\label{sec:occlusion_representation}

HC-MTS constructs $\mathcal F_t$ and $\mathcal O_t$ with a geometry-based visibility procedure. Line-of-sight rays are cast within the field of regard, and the oriented footprints of observed road users are treated as occluders. The resulting partition is used consistently across all frames in the selected history window.

Current hidden-agent proposals are restricted to non-visible, semantically valid support. A spatial seed is
\begin{equation}
    s=(\mathbf p_0,\psi_0,c),
    \label{eq:spatial_seed}
\end{equation}
Here, $\mathbf p_0\in\mathbb R^2$ is the current position and $\psi_0$ is the heading. The class variable $c$ denotes either a vehicle or a pedestrian. Vehicle centers are sampled from lane or drivable corridors, with headings induced by the local lane direction. Pedestrian centers are sampled from crosswalk, walkable, and road-edge support. A seed is retained only if its class-specific footprint is hidden under the sampled-footprint visibility test and does not overlap an observed road user. This produces
\begin{equation}
    \mathcal S_{\mathrm{base}}
    =
    \left\{s=(\mathbf p_0,\psi_0,c)\right\},
    \label{eq:base_seed_set}
\end{equation}
which describes geometrically plausible current positions and headings but does not yet certify temporal feasibility.

\subsection{HC-MTS Stage~I: History-Conditioned Hidden-Mode Inference}
\label{sec:history_inference}

Stage~I of HC-MTS converts the spatial seeds in $\mathcal S_{\mathrm{base}}$ into finite hidden-state modes with explicit historical witnesses.

\paragraph{Finite mode construction.}
A current spatial seed may admit multiple kinematic histories, including straight motion, acceleration, braking, turning, and piecewise control. HC-MTS therefore represents sampled state--history combinations explicitly rather than assigning a continuous speed interval to each seed. A mode is
\begin{equation}
    m
    =
    \left(z_0,\mathbf u^{-},\boldsymbol\tau^{-}\right),
    \qquad
    z_0=(s,v_0),
    \label{eq:mode_definition}
\end{equation}
where $v_0$ is the sampled speed at $t_0$, $\mathbf u^{-}$ is a sampled historical control sequence, and $\boldsymbol\tau^{-}$ is the corresponding backward state trace. This explicit representation is necessary because feasibility of one speed--control combination does not imply feasibility of nearby combinations under the visibility, occupancy, map, and dynamics constraints.

For vehicles, the backward-control sequence is
\begin{equation}
    \mathbf u^-
    =
    \left\{(a_{t_0-j},\kappa_{t_0-j})\right\}_{j=1}^{K_{\max}},
    \label{eq:historical_controls}
\end{equation}
where $a_t$ and $\kappa_t$ denote longitudinal acceleration and path curvature. Constant and piecewise-constant templates represent straight, accelerating, braking, and turning histories. Pedestrian modes use class-specific speed and direction hypotheses without the vehicle curvature model.

\paragraph{Backward witness propagation.}
For a vehicle state $(\mathbf p_t,\psi_t,v_t)$ and control $(a_t,\kappa_t)$, HC-MTS applies the sampled backward update
\begin{align}
    v_{t-1}
    &=v_t-a_t\Delta t,
    &
    \bar v_t
    &=\tfrac12(v_t+v_{t-1}),\\
    \psi_{t-1}
    &=\psi_t-\kappa_t\bar v_t\Delta t,
    &
    \bar\psi_t
    &=\tfrac12(\psi_t+\psi_{t-1}),\\
    \mathbf p_{t-1}
    &=\mathbf p_t-\bar v_t
    \begin{bmatrix}
        \cos\bar\psi_t\\
        \sin\bar\psi_t
    \end{bmatrix}\Delta t.
    \label{eq:backward_dynamics}
\end{align}
Backward propagation terminates at the first contradiction with visibility, observed occupancy, semantic-map support, or class-specific dynamics. The longest valid prefix is denoted by $K^\star(m)$, and the corresponding trace is retained as the backward witness for mode $m$.

\paragraph{History-consistency constraints.}
At every frame required by the selected memory horizon, a mode must satisfy
\begin{align}
    Q(B_t(m))\cap\mathcal F_t
    &=\varnothing,
    &&\text{visibility},
    \label{eq:visibility_constraint}\\
    \operatorname{Footprint}(m_t)\cap\mathcal B_t^{\mathrm{obs}}
    &=\varnothing,
    &&\text{observed occupancy},
    \label{eq:occupancy_constraint}\\
    \operatorname{Footprint}(m_t)
    &\subseteq\mathcal D_c,
    &&\text{semantic map},
    \label{eq:map_constraint}
\end{align}
where $B_t(m)$ is the oriented participant footprint, $Q(B_t(m))$ is its finite visibility-sampling set, $\mathcal B_t^{\mathrm{obs}}$ is the union of observed road-user footprints, and $\mathcal D_c$ is the semantic support for class $c$. Vehicle modes additionally satisfy
\begin{equation}
    v_t\in[0,v_{\max}^{\mathrm{veh}}],
    \qquad
    |\kappa_t v_t|\leq\omega_{\max}^{\mathrm{veh}},
    \qquad
    |\kappa_t v_t^2|\leq a_{\mathrm{lat,max}}^{\mathrm{veh}},
    \label{eq:vehicle_constraints}
\end{equation}
while pedestrian modes satisfy $v_t\in[0,v_{\max}^{\mathrm{ped}}]$ and $\mathbf p_t\in\mathcal D_{\mathrm{ped}}^{\mathrm{hist}}$.

The combined $K$-frame certificate is
\begin{equation}
    \mathcal C_K(m)
    =
    C_{\mathrm{cur}}^{t_0}(m)
    \prod_{j=1}^{K}
    C_{\mathrm{vis}}^{t_0-j}(m)
    C_{\mathrm{occ}}^{t_0-j}(m)
    C_{\mathrm{map}}^{t_0-j}(m)
    C_{\mathrm{dyn}}^{t_0-j}(m).
    \label{eq:consistency}
\end{equation}
Because every requested memory horizon is evaluated from the same sampled modes and the same prefix-validity record,
\begin{equation}
    \mathcal M_{K_2}\subseteq\mathcal M_{K_1},
    \qquad K_2>K_1.
    \label{eq:nested}
\end{equation}
Longer visibility memory can therefore preserve or eliminate an existing sampled hypothesis, but it cannot introduce a new one.

\subsection{HC-MTS Stage~II: Iterative Mini-max Trajectory Search}
\label{sec:counterexample_search}

Stage~II of HC-MTS (Algorithm~\ref{alg:counterexample}) is implemented as a sequence of finite minimax rounds. It is therefore not a single evaluation over a fixed attacker set. Each round starts from the ego trajectory accepted in the preceding round, searches for hidden trajectories that challenge that trajectory, optimizes the ego response to each candidate, and continues only when the selected attacker further reduces the ego vehicle's optimized driving score.

Let $\boldsymbol\tau_i^E$ denote the ego trajectory at the beginning of round $i$, with $\boldsymbol\tau_0^E$ equal to the nominal ego plan. Let $\mathcal P$ be the finite library of candidate attacker motion patterns. For every certified mode $m\in\mathcal M_K$ and pattern $p\in\mathcal P$, HC-MTS searches a bounded set of pattern parameters to instantiate a forward hidden trajectory. Every generated trajectory must satisfy the class-specific dynamics, footprint, semantic-map, observed-occupancy, known-agent, and traffic-legality constraints. The attacker search is collision-seeking: whenever feasible within the finite search budget, the pattern parameters are chosen so that the resulting hidden trajectory collides with the current ego trajectory $\boldsymbol\tau_i^E$. The legal collision-producing candidates found in round $i$ form
\begin{equation}
    \widehat{\mathcal A}_i
    =
    \left\{
        \boldsymbol\tau^A
        \;\middle|\;
        \begin{array}{l}
        \boldsymbol\tau^A \text{ is generated from some }
        m\in\mathcal M_K \text{ and } p\in\mathcal P,\\
        \mathcal C_{\mathrm{fwd}}(\boldsymbol\tau^A)=1,\quad
        \operatorname{Coll}(\boldsymbol\tau_i^E,\boldsymbol\tau^A)=1
        \end{array}
    \right\},
    \label{eq:round_attacker_set}
\end{equation}
where $\mathcal C_{\mathrm{fwd}}$ collects the forward feasibility and legality constraints. Because $\widehat{\mathcal A}_i$ is produced by a bounded search over finite patterns, an empty set means that HC-MTS did not discover a legal collision-producing attacker in that round; it does not prove that none exists in the continuous trajectory space.

\paragraph{Inner ego-response optimization.}
For every candidate $\boldsymbol\tau^A\in\widehat{\mathcal A}_i$, HC-MTS independently optimizes an ego response. Let $\widehat{\mathcal R}_{E,i}(\boldsymbol\tau^A)$ be the finite response set searched against that candidate. The optimized ego value is
\begin{equation}
    V_i(\boldsymbol\tau^A)
    =
    \max_{\boldsymbol\tau^E\in
    \widehat{\mathcal R}_{E,i}(\boldsymbol\tau^A)}
    J(\boldsymbol\tau^E,\boldsymbol\tau^A),
    \label{eq:round_best_response_value}
\end{equation}
with corresponding best response
\begin{equation}
    \boldsymbol\tau_i^{E*}(\boldsymbol\tau^A)
    \in
    \arg\max_{\boldsymbol\tau^E\in
    \widehat{\mathcal R}_{E,i}(\boldsymbol\tau^A)}
    J(\boldsymbol\tau^E,\boldsymbol\tau^A).
    \label{eq:round_best_response}
\end{equation}
Here, larger $J$ indicates better ego performance. Collision receives strict priority, while collision-free responses are ranked by the destination-attainment and ride-comfort terms defined in the driving objective. 

\paragraph{Within-round minimization.}
After solving the inner response problem for every candidate, HC-MTS selects the attacker that produces the lowest optimized ego score:
\begin{align}
    \boldsymbol\tau_i^{A*}
    &\in
    \arg\min_{\boldsymbol\tau^A\in\widehat{\mathcal A}_i}
    V_i(\boldsymbol\tau^A),
    \label{eq:round_attacker_selection}\\
    S_i
    &=
    V_i(\boldsymbol\tau_i^{A*})
    =
    J\!\left(
        \boldsymbol\tau_i^{E*}(\boldsymbol\tau_i^{A*}),
        \boldsymbol\tau_i^{A*}
    \right).
    \label{eq:round_score}
\end{align}
Equations~\eqref{eq:round_best_response_value}--\eqref{eq:round_score} define the mini-max problem solved within round $i$: the inner maximization finds the strongest ego response to each fixed attacker, and the outer minimization selects the attacker that leaves the ego with the lowest best-response score.

\begin{algorithm}[!htbp]
\caption{HC-MTS Stage~II: Iterative mini-max trajectory search}
\label{alg:counterexample}
\begin{algorithmic}[1]
\Require Certified modes $\mathcal M_K$, attacker-pattern library $\mathcal P$, nominal ego plan $\boldsymbol\tau_0^E$, score tolerance $\epsilon$, and round budget $L$
\Ensure Last accepted attacker--response pair and its optimized ego score
\State $S_{\mathrm{prev}}\gets+\infty$; accepted pair $\gets\varnothing$
\For{$i=0,\ldots,L-1$}
    \State Generate a finite set $\widehat{\mathcal A}_i$ of legal, collision-seeking attacker trajectories from $\mathcal M_K$ and $\mathcal P$ against $\boldsymbol\tau_i^E$
    \If{$\widehat{\mathcal A}_i=\varnothing$}
        \State \textbf{break}
    \EndIf
    \For{each $\boldsymbol\tau^A\in\widehat{\mathcal A}_i$}
        \State Optimize the ego response $\boldsymbol\tau_i^{E*}(\boldsymbol\tau^A)$
        \State $V_i(\boldsymbol\tau^A)\gets J\!\left(\boldsymbol\tau_i^{E*}(\boldsymbol\tau^A),\boldsymbol\tau^A\right)$
    \EndFor
    \State Optionally refine low-value attacker patterns; revalidate each refined trajectory and re-optimize its ego response
    \State $\boldsymbol\tau_i^{A*}\gets\arg\min_{\boldsymbol\tau^A\in\widehat{\mathcal A}_i}V_i(\boldsymbol\tau^A)$
    \State $\boldsymbol\tau_i^{E*}\gets\boldsymbol\tau_i^{E*}(\boldsymbol\tau_i^{A*})$; $S_i\gets V_i(\boldsymbol\tau_i^{A*})$
    \If{$i>0$ \textbf{and} $S_i\geq S_{\mathrm{prev}}-\epsilon$}
        \State \textbf{break} \Comment{the optimized ego score no longer decreases}
    \EndIf
    \State Accept $(\boldsymbol\tau_i^{A*},\boldsymbol\tau_i^{E*},S_i)$; $S_{\mathrm{prev}}\gets S_i$
    \If{$\boldsymbol\tau_i^{E*}$ is not collision-free}
        \State Mark the accepted pair as unresolved and \textbf{break}
    \EndIf
    \State $\boldsymbol\tau_{i+1}^E\gets\boldsymbol\tau_i^{E*}$
\EndFor
\State \Return the last accepted pair and $S_{\mathrm{prev}}$
\end{algorithmic}
\end{algorithm}

\paragraph{Cross-round descent and termination.}
The selected pair is accepted only when it produces a sufficient decrease relative to the preceding accepted round. For $i\geq 1$, the acceptance condition is
\begin{equation}
    S_i < S_{i-1}-\epsilon,
    \label{eq:round_descent}
\end{equation}
where $\epsilon\geq0$ is the score-decrease tolerance. If Eq.~\eqref{eq:round_descent} holds, HC-MTS accepts
$\bigl(\boldsymbol\tau_i^{A*},
\boldsymbol\tau_i^{E*}(\boldsymbol\tau_i^{A*})\bigr)$ and uses the optimized ego response as the plan for the next round:
\begin{equation}
    \boldsymbol\tau_{i+1}^E
    =
    \boldsymbol\tau_i^{E*}(\boldsymbol\tau_i^{A*}).
    \label{eq:ego_round_update}
\end{equation}
The next attacker search then attempts to reduce the ego score again by finding a new trajectory against this updated plan. If $S_i\geq S_{i-1}-\epsilon$, the current round does not improve the adversarial objective, the pair is not accepted, and HC-MTS terminates with the last accepted pair. 


HC-MTS also terminates when no legal collision-producing attacker is discovered, when the round budget is exhausted, or when the selected attacker admits no collision-free response. The last case is reported as an unresolved counterexample because no collision-free ego trajectory is available to initialize another round.

\paragraph{Finite candidate search and refinement.}
Within each round, HC-MTS evaluates only a finite number of attacker trajectories generated from the candidate patterns. A cluster-balanced shortlist may be used to distribute this budget across distinct current hidden modes or spatial seeds. Candidate patterns associated with low best-response scores can be refined by one round of clipped Gaussian elite sampling inspired by the cross-entropy method \cite{deboer2005cem}. Every refined attacker is revalidated and, crucially, the inner ego-response optimization is rerun for that refined trajectory before its score is compared with other candidates. The refinement, therefore, seeks a lower \emph{optimized} ego score rather than exploiting a response computed for a different attacker.

\section{Numerical Results}
\subsection{Experiment Setups}
\paragraph{Research questions.}
The experiments are designed to answer two research questions:
\textbf{RQ1:} How does the length of the visibility history affect the
set of hidden states that remain consistent with the available
observations?
\textbf{RQ2:} Can HC-MTS discover legal, history-consistent hidden
trajectories that reduce the ego vehicle's best achievable driving
score after response optimization?
The first question evaluates the history-conditioned inference stage,
whereas the second evaluates the iterative response-aware minimax
search.

\paragraph{Dataset and scenario selection.}
We evaluate HC-MTS on eight vectorized scenarios from the WOMD \cite{ettinger2021womd}. The scenarios are
randomly sampled from the subset for which the geometric visibility
procedure produces a nonempty current blind region relevant to the ego
vehicle's planning horizon. The scenarios are selected once before
inspecting the outputs of HC-MTS and are retained regardless of the
resulting hidden-area reduction or counterexample-search outcome. The
same eight scenarios are used in all experiments. For each scene, the
method uses the vectorized semantic map, the ego state, and the observed
road-user tracks up to the current time $t_0$. No future state of a
hypothetical hidden participant is used for either history certification
or adversarial trajectory search. All scene elements are represented in
the ego-fixed coordinate system defined in
Section~\ref{sec:occlusion_representation}.

\subsection{Effect of Visibility Memory}
\label{sec:k_ablation_results}


Table~\ref{table:2} reports the number of
history-consistent hidden seeds retained under different
visibility-memory lengths. Increasing $K$ consistently reduces the
retained hypothesis set. From $K=1$ to $K=20$, the mean per-scenario
vehicle, pedestrian, and total seed counts decrease by $18.12\%$,
$21.67\%$, and $18.45\%$, respectively. The larger reduction for
pedestrians is consistent with their smaller backward-reachable
displacement, which makes their historical trajectories more likely to
conflict with previously visible-free regions. In contrast, vehicle
hypotheses can remain feasible by originating farther along
continuously occluded road corridors.

\begin{table}[H]
    \centering
    \caption{
        Average number of retained history-consistent hidden seeds over
        eight WOMD scenarios. Reduction is the mean per-scene percentage
        decrease relative to $K=1$. Total denotes all retained vehicle
        and pedestrian seeds.
    }
    \label{tab:k_seed_count_ablation}
    \setlength{\tabcolsep}{3.5pt}
    \resizebox{0.8\columnwidth}{!}{
    \begin{tabular}{c rr rr rr}
        \toprule
        & \multicolumn{2}{c}{Vehicle}
        & \multicolumn{2}{c}{Pedestrian}
        & \multicolumn{2}{c}{Total} \\
        \cmidrule(lr){2-3}
        \cmidrule(lr){4-5}
        \cmidrule(lr){6-7}
        $K$
        & Count & Reduction
        & Count & Reduction
        & Count & Reduction \\
        \midrule
        1  & 3392.12 & 0.00\%
           & 342.62  & 0.00\%
           & 3734.75 & 0.00\% \\
        5  & 3155.12 & 6.99\%
           & 311.12  & 9.19\%
           & 3466.25 & 7.19\% \\
        10 & 2992.62 & 11.78\%
           & 289.25  & 15.58\%
           & 3281.88 & 12.13\% \\
        20 & 2777.38 & 18.12\%
           & 268.38  & 21.67\%
           & 3045.75 & 18.45\% \\
        \bottomrule
    \label{table:2}
    \end{tabular}}
\end{table}

\begin{figure}[H]
    \centering
    \includegraphics[width=1\linewidth]{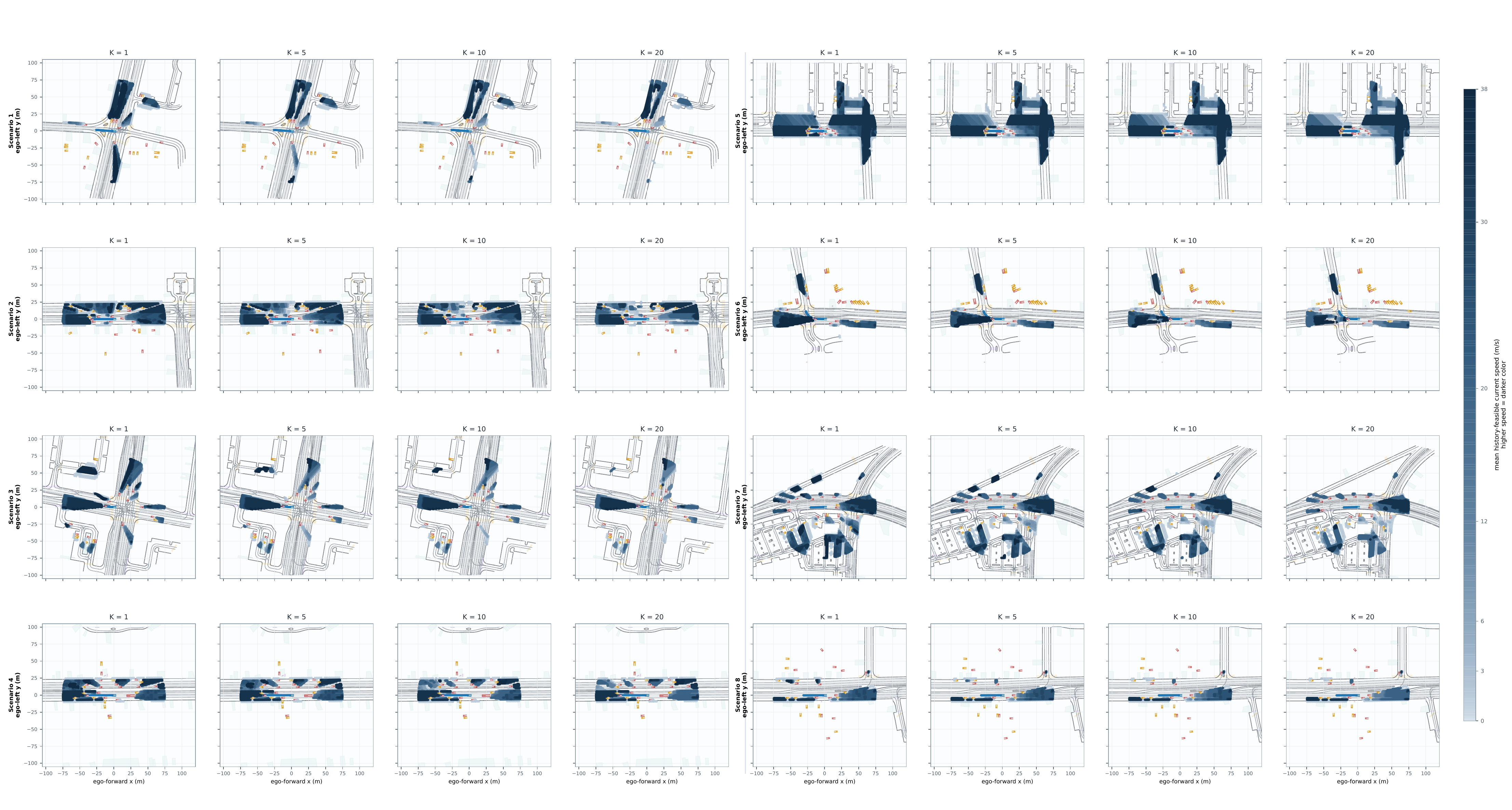}
    \caption{Comparison of hidden-seed distributions using different temporal history lengths K across eight WOMD scenarios. Marker color indicates the mean history-feasible current speed, with darker markers representing higher speeds.  }
    \label{fig:2}
\end{figure}

Fig.~\ref{fig:2} provides the corresponding qualitative comparison. Longer histories remove spatially coherent seed clusters rather than uniformly reducing the entire current occlusion mask. Scenarios 1 and 3 show substantial contraction, indicating that ego motion reveals large portions of the candidate hidden corridors. Scenarios 5 and 8 change less because their occlusions remain persistent over the evaluated history. Accordingly, the $K=1$ to $K=20$ reduction varies from $4.57\%$ to $35.44\%$ across the eight scenarios, showing that the value of temporal visibility depends strongly on ego motion, occluder geometry, and road topology.

In Fig.~\ref{fig:2}, darker markers represent higher mean history-feasible current speeds. The color is used only for visualization: each seed may retain multiple feasible motion modes, and the complete mode set, rather than the displayed mean speed, is passed to the subsequent trajectory optimization. The persistence of several high-speed seeds in continuously occluded corridors also shows that the method does not remove candidates based on danger alone. Seeds are pruned only when their historical existence is contradicted by visibility, dynamics, or physical occupancy constraints.

\subsection{Response-Aware Critical-Trajectory Search}

\begin{figure}[H]
    \centering
    \includegraphics[width=1\linewidth]{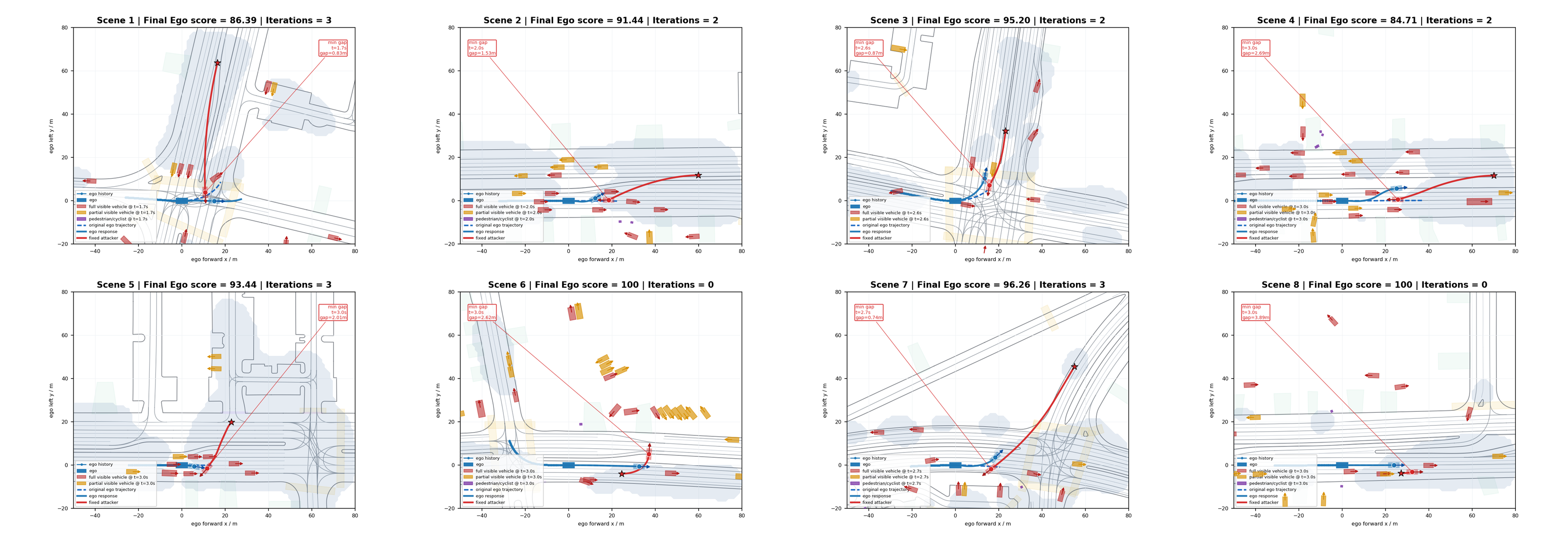}
    \caption{Response-aware hidden-attacker search across eight WOMD scenes. Each panel shows the final result for one scene and reports the ego score and the number of outer best-response iterations. The dashed blue trajectory is the nominal ego trajectory, the solid blue trajectory is the optimized ego response, and the red trajectory is the selected hidden attacker. Vehicle footprints are displayed at the corresponding evaluation time. A finite ego score below $100$ indicates that a legal collision-producing attacker was found, but the ego response oracle could still avoid the collision at the cost of reduced progress, and route tracking. A score of $100$ indicates that no legal collision-producing attacker was identified under the current search budget. A score of $-\infty$ indicates that no collision-free ego response could be found against the selected attacker (not found within the finite search budget).}
    \label{fig:3}
\end{figure}

In Fig.~\ref{fig:3}, the eight scenes exhibit two of the three possible outcomes. In
Scenes~1--5 and~7, HC-MTS finds a legal, history-consistent attacker
that collides with the current ego trajectory, while the finite
ego-response oracle still returns a collision-free response. The
corresponding final ego scores are $86.39$, $91.44$, $95.20$, $84.71$,
$93.44$, and $96.26$, respectively, with a mean of $91.24$.
Scene~4 shows the largest degradation, with a score of $84.71$.
These results show that collision with the reference ego trajectory
alone can overstate attacker severity: all six attackers remain
avoidable, although replanning incurs losses in destination attainment, or ride comfort.

In Scenes~6 and~8, no legal collision-producing attacker is found
within the finite search budget; consequently, the ego score remains
at the nominal value of $100$, and the outer game terminates without
an update. No scene yields a score of $-\infty$, meaning that the
finite oracle finds at least one collision-free response for every
attacker identified in this evaluation. Current outcomes are
budget-dependent and do not constitute formal guarantees over the
continuous hidden-state and control spaces. Overall, the results
highlight the value of response-aware evaluation for distinguishing
avoidable reference-trajectory collisions from threats that remain
unresolved after ego replanning.

\section{Conclusion}
\label{sec:conclusion}

This work introduced \emph{History-Conditioned Minimax Trajectory Search} (HC-MTS), a unified framework for identifying safety-critical trajectories of traffic participants concealed by occlusion. HC-MTS couples history-consistent hidden-state inference with response-aware adversarial search, enabling candidate trajectories to be evaluated not only for physical and observational feasibility, but also for their effect on the ego vehicle’s best achievable task-level performance.

The experimental results demonstrate two key advantages of the proposed formulation. First, conditioning on multi-frame visibility and occupancy evidence eliminates hidden-state hypotheses that are compatible with the current occluded region but inconsistent with the observation history. Second, optimizing the ego response before assessing trajectory criticality yields a more principled evaluation than measuring collision risk against a fixed nominal plan. This response-aware criterion distinguishes hazards that can be mitigated through replanning from those that remain safety-critical under the available ego strategy, thereby providing a stronger and more informative basis for evaluating autonomous-driving planners.

The current limitation of HC-MTS is its reliance on a heuristic solution algorithm, which does not guarantee global optimality and may become computationally demanding as the scenario complexity and search space increase. Future work will therefore focus on two directions. First, we will evaluate HC-MTS on larger-scale datasets and a broader range of occlusion-critical driving scenarios to assess its robustness and generalizability. Second, we will refine the solution algorithm to improve computational efficiency, optimization stability, and solution quality, and to reduce its dependence on heuristic search.

\section{Acknowledgment}
This work is based upon the work supported by the National Center for Transportation Cybersecurity and Resiliency (TraCR) (a U.S. Department of Transportation National University Transportation Center) headquartered at Clemson University, Clemson, South Carolina, USA. Any opinions, findings, conclusions, and recommendations expressed in this material are those of the author(s) and do not necessarily reflect the views of TraCR, and the U.S. Government assumes no liability for the contents or use thereof.

\bibliographystyle{splncs04}
\bibliography{egbib}
\end{document}